\documentclass[a4paper,fleqn]{cas-dc}

\usepackage[numbers]{natbib}

\usepackage{color,soul}
\usepackage{url}
\usepackage{algorithm}
\usepackage{algorithmic}
\usepackage{amsmath}
\usepackage{subfigure}
\usepackage{setspace}
\usepackage{amssymb}
\usepackage{pifont}
\usepackage{enumitem}

\newtheorem{Problem}{Problem Definition}

\def\tsc#1{\csdef{#1}{\textsc{\lowercase{#1}}\xspace}}
\tsc{WGM}
\tsc{QE}
\tsc{EP}
\tsc{PMS}
\tsc{BEC}
\tsc{DE}
\begin{document}
\let\WriteBookmarks\relax
\def\floatpagepagefraction{1}
\def\textpagefraction{.001}
\shorttitle{KnowAugNet}
\shortauthors{Yang An et~al.}

\title [mode = title]{KnowAugNet: Multi-source medical knowledge augmented medication prediction network with multi-Level graph contrastive learning}

\author[1]{Yang An}
\credit{Conceptualization, Methodology, Software, Validation, Writing-original draft, Formal Analysis, Data Curation}
\address[1]{School of Software, North University of China, Taiyuan, China, 030051}

\author[2]{Bo Jin}[orcid=0000-0002-4094-7499]
\cormark[1]
\ead{jinbo@dlut.edu.cn}
\credit{Supervision, Funding acquisition}
\address[2]{School of Innovation and Entrepreneurship, Dalian University of Technology, Dalian, China, 116024}

\author[3]{Xiaopeng Wei}
\cormark[1]
\ead{xpwei@dlut.edu.cn}
\credit{Project administration}
\address[3]{School of Computer Science and Technology, Dalian University of Technology, Dalian, China, 116024}


\begin{abstract}
Predicting medications is a crucial task in many intelligent healthcare systems. It can assist doctors in making informed medication decisions for patients according to electronic medical records (EMRs).
However, medication prediction is a challenging data mining task due to the complex relations between medical codes. Most existing studies focus on utilizing inherent relations between homogeneous codes of medical ontology graph to enhance their representations using supervised methods, and few studies pay attention to the valuable relations between heterogeneous or homogeneous medical codes, e.g., synergistic relations between medications, concurrent relations between diseases, therapeutic relations between medication and disease from history EMRs, which further limits the prediction performance and application scenarios. 
Therefore, to address these limitations, this paper proposes \textbf{KnowAugNet}, a multi-sourced medical knowledge augmented medication prediction network which can fully capture the diverse relations between medical codes via multi-level graph contrastive learning framework. Specifically, KnowAugNet first leverages the graph contrastive learning using graph attention network as the encoder to capture the implicit relations between homogeneous medical codes from the medical ontology graph and obtains the knowledge augmented medical codes embedding vectors. Then, it utilizes the graph contrastive learning using a weighted graph convolutional network as the encoder to capture the correlative relations between homogeneous or heterogeneous medical codes from the constructed medical prior relation graph and obtains the relation augmented medical codes embedding vectors. Finally, the augmented medical codes embedding vectors and the supervised medical codes embedding vectors are retrieved and input to the sequential learning network to capture the temporal relations of medical codes and predict medications for patients. The experimental results demonstrate a consistent superiority of our model over several baseline models in the medication prediction task.

\end{abstract}


\maketitle
\section{Introduction}
\label{Introduction}
The availability of immense accumulation of electronic medical records (EMRs) data and the advancements of deep computational methods laid a solid foundation for intelligent healthcare applications, such as disease risk prediction \cite{Ye2020LSANML,Choi2018MiMEMM,Zhang2019ATTAINAT} and medication prediction task \cite{Shang2019GAMENetGA,shang2019pre,He2020AttentionAM}.
Among them, predicting medications for patients can assist doctors in making clinical decisions more efficiently, which further enables clinicians to devote more time to the disease communication stage in case of misdiagnosis. Thus, it will be conducive to improving the medical service quality. 
In this case, many deep learning-based medication prediction models emerge as the times require.
However, most of them fail to meet such a scenario that several medical experts conduct joint consultations for patients, in which the medical knowledge such as the common-sense medical ontology and the empirical medical knowledge from the history EMRs data are considered in the medications decision-making process, and thus result in suboptimal performance.
Therefore, how to effectively capturing the complex and diverse relations between medical codes from multi-source medical knowledge to augment medication prediction is a highly challenging task but meaningful.

\begin{figure}
\centering
    \includegraphics[width=1\linewidth]{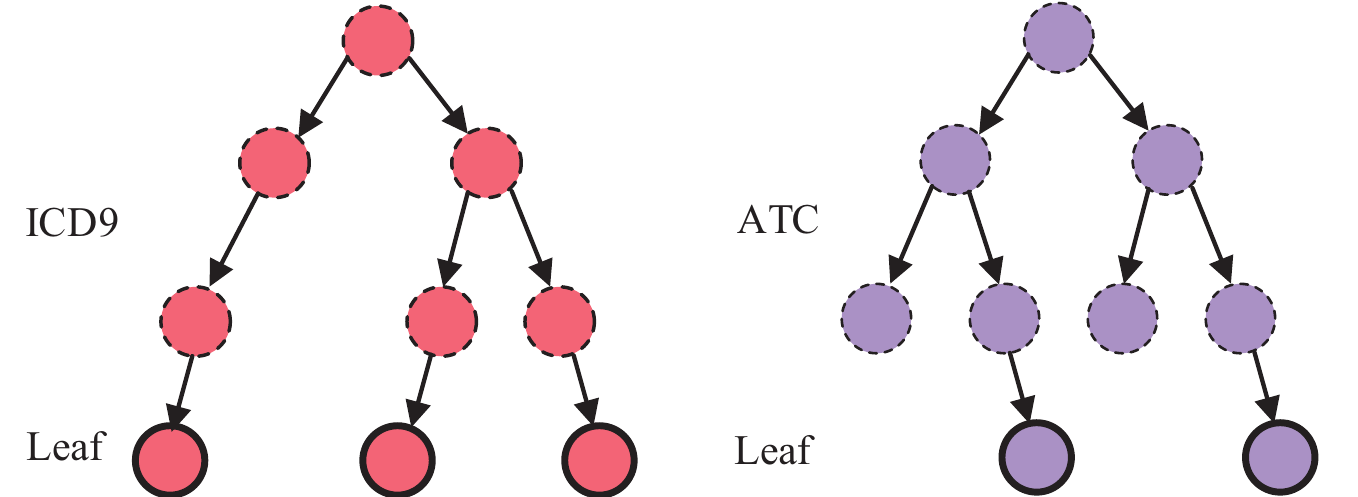}
    \caption{medical ontology graph, left: diagnosis code ontology graph (ICD 9), right: medication code ontology graph (ATC).}
    \vspace{-0.2cm}
    \label{fig:Chap_HG}
\end{figure}

As illustrated in \ref{fig:Chap_HG}, the hierarchical structures implicit in the diagnosis ontology graph and medication ontology graph (a kind of common-sense medical domain knowledge) imply the inherent relations between homogeneous medical codes, which will contribute to the representation learning. The existing studies, such as GRAM \cite{Choi2017GRAMGA} and KAME \cite{Ma2018KAMEKA} utilize the diagnosis ontology graph to enhance the representations of diagnosis codes by infusing the information from relational medical codes in a supervised method. While Shang et al. \cite{shang2019pre} proposed G-BIRT to combine the graph neural network and bidirectional encoder representation from transformers (BERT) to enhance medical code representations in a pre-training method. For medical code representation augmentation, the above models consider the inherent relations between homogeneous medical codes in medical domain ontology graphs in a supervised or self-supervised way, which result in that the learned medical codes representations assisted by the ontology graph can not be easily transferred for different predictive tasks such as from diagnosis prediction to medication prediction despite using the same dataset. This phenomenon will cause repetitive model training for obtaining medical code representations when applied in every different downstream task. When the downstream task is unknown, the purely supervised methods will not work for obtaining the medical codes representations in case of need.
Therefore, there is an urgent need to develop a new method for learning the embeddings of medical codes based on the medical ontology graph in an unsupervised way, which could further contribute to the downstream tasks in different clinical scenarios.

Additionally, the models mentioned above capture the inherent relations between homogeneous medical codes of the hierarchical structures in medical ontology graphs. At the same time, they all neglect the valuable correlative relations between heterogeneous or homogeneous medical codes implicit in historical EMRs data, which is usually treated as a kind of empirical medical knowledge. For example, clinicians tend to simultaneously utilize several medications for the patients to enhance the therapeutic effect, which reflects the synergistic relations between medications. Major diseases are often accompanied by inevitable concurrent diseases or symptoms, which reflect the concurrent relations between diseases. Moreover, in a prescription, the medications are prescribed for the corresponding diseases or symptoms, reflecting the therapeutic causal relations between medication and disease. Few studies can explicitly represent and capture the above meaningful relations hidden in EMRs data for medication prediction.

To tackle the aforementioned limitations and further improve the performance of medication prediction, we propose a multi-sourced medical knowledge augmented network with multi-level graph contrastive learning, which aims to capture the diverse relations between homogeneous or heterogeneous medical codes for enhancing their representations.
Firstly, similar to the existing method G-BIRT \cite{shang2019pre}, the inherent common-sense medical ontology graph (a kind of medical domain knowledge, shown in Figure \ref{fig:Chap_HG}) will be taken into consideration for capturing the inherent local relations.
Meanwhile, considering the correlative relations between heterogeneous or homogeneous medical codes from historical EMRs data, and given the advantages of graph representation learning \cite{xu2018how}, we construct a medical prior relation graph (shown in Figure \ref{fig:Chap_RG}) based on the co-occurrence of medical codes including diagnosis code and medication code in a single visit, and further capture the implicit global relation.
Secondly, for the problems of label dependency and repetitive model training in the process of existing ontology graph representation learning methods, considering the superiority of graph contrastive learning that can provide better representations for downstream tasks without supervised labels and can facilitate to improve the model robustness \cite{NEURIPS2019_a2b15837}, we incorporate the improved graph contrastive learning framework, based on DGI\cite{velickovic2018deep}, to realize the unsupervised representations learning of medical codes from multi-source knowledge graph.
In this way, the medical codes representations could be mutually augmented via infusing the information from relational medical codes through capturing the inherent local relations between medical codes of the ontology graph and the implicit global relations between medical codes of the medical prior relation graph.
Finally, the retrieved augmented medical codes representations according to the medical codes are input into the supervised sequential learning model to capture the temporal relations between medical codes. Then the obtained comprehensive patient representation could be leveraged to predict the medications to assist clinical decision-making.

Thus, the related technical contributions of this paper can be summarized as follows:
\begin{itemize}[leftmargin=*]
    \item We propose a novel multi-source medical knowledge augmented medication prediction network, called \textbf{KnowAugNet}, with a multi-level graph contrastive learning framework. To the best of our knowledge, our model is the first to capture the valuable relations between medical codes, which would be further used to augment the medical codes representations via a cascaded unsupervised method on the inherent ontology graphs and constructed medical prior relation graph.
    \item We incorporate different graph encoders, including the graph attention network and weighted graph convolutional network, into the multi-level graph contrastive learning networks for considering the meaningful relation weights between different medical codes.
    \item We present a sequential learning network to integrate the multi-source embedding vectors of medical codes into patient representation, by which the temporal relations between medical codes are also captured, for predicting the medications for patients.
    \item The presented model is validated using the real-world dataset and compared with those of the state-of-the-art models. The results show that the proposed model exhibits relatively competitive performance.
\end{itemize}

\section{Related works}
\label{sec:relatedWorks}
\subsection{Medication prediction}

With the accumulation of EMRs data and the development of computational models, especially the deep learning models, medication prediction, one of the predictive applications in intelligent healthcare systems, has been studied by many researchers owing to its practical significance. 
Shang et al. \cite{Shang2019GAMENetGA} have categorized the medication prediction algorithms into instance-based and longitudinal sequential prediction methods.
Instance-based medication prediction methods mainly focus on capturing the nonlinear relations between diagnosed disease status and the output prescribed medications. 
For instance, Zhang et al. \cite{Zhang2017LEAPLT} formulates the medication problem as a sequential decision making problem and uses the recurrent neural networks to encode the label dependency.
Wang et al. \cite{Wang2018PersonalizedPF} introduce three linear models to concatenate the multi-source patient information, such as demographic information, laboratory indicators information and diagnosis outcomes, for personalized medication prediction.
While Wang et al. \cite{Wang2019OrderfreeMC} transform the medication prediction task into an unordered Markov decision process for predicting the prescription medications step by step.
The critical temporal information contained in historical EMR data is overlooked, which results in poor prediction performance.

Therefore, the longitudinal sequential prediction models considering the temporal relations between historical medical records are prevailing for the medication prediction task. 
Jin et al. \cite{Jin2018KDD} present three different heterogeneous LSTM models to capture the interaction between heterogeneous temporal sequence data and, in turn, incorporates two heterogeneous sequence information into the patient representation to predict the next stage of treatment medications.
Shang et al. \cite{Shang2019GAMENetGA} infuses the sequential information, including diagnosis and procedure information, through multi-channel sequence learning networks for learning comprehensive patient representation for medication prediction. While DMNC\cite{Le2018DualMN} and AMANet \cite{He2020AttentionAM} mainly integrate the attention networks to capture the interactions between procedure and diagnosis sequences and further model the sequential dependency. MeSIN \cite{an2021mesin} not only models the temporal dependency between sequential medical records but also captures the relations between hierarchical sequences for medication prediction problems.
However, they all consider mining the inherent relations between multiple medical sequences of EMRs data and neglect the empirical knowledge implicit in EMRs data and the external common-sense medical knowledge.

\subsection{Graph representation learning}

Graph neural network(GNN) \cite{Scarselli2009TheGN} is an effective graph representation learning framework.
It follows a neighbourhood aggregation mechanism. That is, it calculates the representation vector by recursively aggregating and transforming the representation vectors of its adjacent nodes \cite{Kipf2017SemiSupervisedCW,xu2018how}. 
It can effectively make use of the topological relationship between graph nodes to construct the relation between connected graph nodes and enhance the representation ability of graph nodes.
Because of the powerful ability of GNN, it has been widely used in biological and health informatics to model the valuable relations between multiple entities \cite{Ruiz2021IdentificationOD,Zitnik2018ModelingPS,Shang2019GAMENetGA,ye2021medpath}.
In this way, the learned medical codes embedding vectors in health informatics can be augmented to facilitate the downstream predictive tasks.
For instance, GRAM \cite{Choi2017GRAMGA}, and KAME \cite{Ma2018KAMEKA} utilize the diagnosis ontology graph to enhance the representations of diagnosis codes by infusing the information from relational diagnosis codes from the ontology graph with the help of the attention mechanism. 
Zhang et al. \cite{zhang2019knowrisk} and Ye et al. \cite{ye2021medpath} all introduce the knowledge into the sequential network for enhancing the representation learning and obtaining an interpretable disease risk prediction results via medical knowledge infusion. 
However, Sun et al. \cite{sun2020interpretable} formulates the disease risk prediction task as a random walk along with a knowledge graph (KF) to get the explainable results.
While in the medication prediction domain, 
Mao et al. \cite{mao2022medgcn} utilizes the historical EMRs data (including the lab test, medications, encounters, and patients) to construct the medical graph for incorporating the information of correlative entities into code representation for medication prediction and lab test imputation.
Su et al. \cite{su2020gate} firstly construct the dynamic co-occurrence graph for every patient at each admission record and then present a graph-attention augmented sequential network to model better the inherent structural and temporal information for medication prediction. 
However, the significant relations between medical codes are not fully leveraged owing to the lack of global graph representation learning and just extracting the local sub-graph from the global guidance relation graph.

Although the above GNN based models have achieved relatively good performance in multiple tasks, there are still many problems, such as complex and laborious obtaining the supervised labels and need to re-training when facing new tasks or feature changes. So researchers have sought new training methods for GNN, such as pre-training and self-supervised methods.
The pre-training methods require training on easily accessible large-scale graph data in an unsupervised or coarse-grained supervised way to provide initial parameters for downstream tasks and then fine-tuning on downstream tasks.
For example, Hu et al. \cite{Hu2020GPTGNNGP} presents a generative graph pre-training task, which is to define the conditional probability distribution of the generative graph according to the hypothesis and then takes the recovery of graph structure and attribute as the training task of GNN.
In the field of health informatics, Shang et al. \cite{shang2019pre} and Wang et al. \cite{wang2021medication} combine the graph neural network and bidirectional encoder representation from transformers (BERT) to capture the inherent relations between homogeneous medical codes from medical ontology graphs and enhance the medical code representations in a pre-training method.
However, graph self-supervised learning belongs to an unsupervised graph representation learning approach, which does not depend on labels but only on the topology and node information of the graph itself.
For instance, Petar \cite{velickovic2018deep} incorporates Deep InfoMax \cite{Hjelm2019LearningDR} into the graph learning domain and models a general self-supervised learning framework based on the mutual information maximization. Similarly, Kaveh et al. \cite{pmlr-v119-hassani20a} trains the graph by maximizing the representation graph encoding of different graph structure perspectives.
In biomedicine, Sun et al. \cite{Sun2021MoCLCL} proposes a novel molecular graph contrastive learning framework, which can leverage the local and global domain knowledge to enhance the graph representation learning.
Therefore, inspired by the advantages of contrastive learning methods, we incorporate the graph contrastive learning framework with two different novel graph encoders for the embedding learning of medical code from the medical ontology graph and medical prior relation graph.

\section{Method}
\subsection{Problem formulation}
\label{PD}

\noindent\textbf{Definition 1 (Clinical patient records)}
As the complexity of EMRs data, we firstly define the experimental data from EMRs data using the mathematical symbols to facilitate the latter illustration of our medication prediction model and generalize the relevant dataset in the future.
Specifically, the clinical EMRs data generally contain the longitudinal sequential history medical records, in which we can represent each patient as a sequence of multivariate observations: $\boldsymbol{P} = \left\{\mathcal{X}^1,\mathcal{X}^2,...,\mathcal{X}^{T_n}\right\}$, where $n \in\{1,2, \ldots, N\}$, $N$ is patients' total number, and $T_n\geq 1$ indicates the number of observations for the $n$-th patient.
To reduce clutter and without loss of generality, we describe the model for a patient, and the subscript (n) will be dropped whenever it is unambiguous.
For each observation of a patient, $\mathcal{X}^{t} = [\boldsymbol{c}^t_d,\boldsymbol{c}^t_m]$ represents the medical codes set including diagnosis codes $\boldsymbol{c}^t_d$ and medication codes $\boldsymbol{c}^t_m$ at timestamp $t$. 
For simplicity, $\boldsymbol{c}^t_v$ is used to indicate the unified definition of medical codes. 
For example, $\boldsymbol{c}^j_v \in \{0,1\}^{\left|\mathcal{C}_V\right|}$ is a multi hot vector, where $c^j_v=1$ if the $i$-th medical code of $j$-th observation is included in the subset of medical codes $\mathcal{C} = [\mathcal{C}_d,\mathcal{C}_m]$, and  $\mathcal{C}_d$ and $\mathcal{C}_m$ respectively denotes the union sets of diagnosis codes and medication codes. 

\noindent\textbf{Definition 2 (Medical knowledge graph)}
The incorporated multi-sourced medical knowledge graphs, in this paper, include the common-sense medical ontology graph and the empirical medical prior relation graph.
The former is classified based on the classification system with tree structure possessing the parent-child relations, including the ICD-9 diagnosis code ontology graph $\mathcal{G}_d=\{\mathcal{V}_d,\mathcal{E}_d\}$ and the ATC medication code ontology graph $\mathcal{G}_m=\{\mathcal{V}_m,\mathcal{E}_m\}$, which is called \textbf{Medical ontology graph}.
Note that we employ the unified symbol $\mathcal{G}_* = \{\mathcal{V}_*,\mathcal{E}_*\}$ represent the different medical ontology graphs, where $\mathcal{V}_*$ is the set of graph nodes (medical codes), and $\mathcal{E}_*$ is the set of edges (the relations of medical codes).
Among them, the set of medical codes $\mathcal{C}_*$ constitutes the leaf nodes of medical ontology graphs, and the set of all the graph nodes of $\mathcal{G}_*$ satisfies $\mathcal{V}_*= \mathcal{C}_* \cup \mathcal{C}^{'}$, where $\mathcal{C}^{'}$ denotes the set of non-leaf nodes.
In addition, as illustrated in \ref{fig:Chap_RG}, \textbf{the medical prior relation graph} $\mathcal{G}_r = \{\mathcal{V}_r,\mathcal{E}_r,\mathcal{W}_r\}$ is constructed based on the history medical records, where $\mathcal{V}_r$ represents the graph nodes set, also called the medical codes set $\mathcal{C} = [\mathcal{C}_d,\mathcal{C}_m]$, $\mathcal{E}_r$ is the graph edges set, $\mathcal{W}_r$ represents the relation weight. Moreover, the heterogeneity between diagnosis code and medication code would be neglected for the reason that the relation strength between medical codes is the critical consideration factor in this paper.

\begin{Problem}[\textbf{Medication prediction}]
Given the history electronic medical records of a patient $\left\{\mathcal{X}^{1},\mathcal{X}^{2},...,\mathcal{X}^{t-1}\right\}$, the current timestamp $t$-th diagnosis codes $\mathcal{C}_d^{t}$, the constructed medical ontology graph $\mathcal{G}_*$ and the medical prior relation graph $\mathcal{G}_r$,
our goal is to fully utilize above multi-sourced related information to predict reasonable medications for patient by generating the multi-label output $\hat{\boldsymbol{y}}^m_{t} \in \{0,1\}^{\left|\mathcal{C}_{m}\right|}$:
\vspace{-0.3cm}
\begin{equation}
    \hat{\boldsymbol{y}}^m_{t} = f(\left\{\mathcal{X}^{1},\mathcal{X}^{2},...,\mathcal{X}^{t-1}\right\}, \boldsymbol{c}^t_d).
\label{Chap_target}
\end{equation}
\end{Problem}

\subsection{Proposed KnowAugNet}
\label{sec:Chap_Methods}
\begin{figure}
\centering
    \includegraphics[width=1\linewidth]{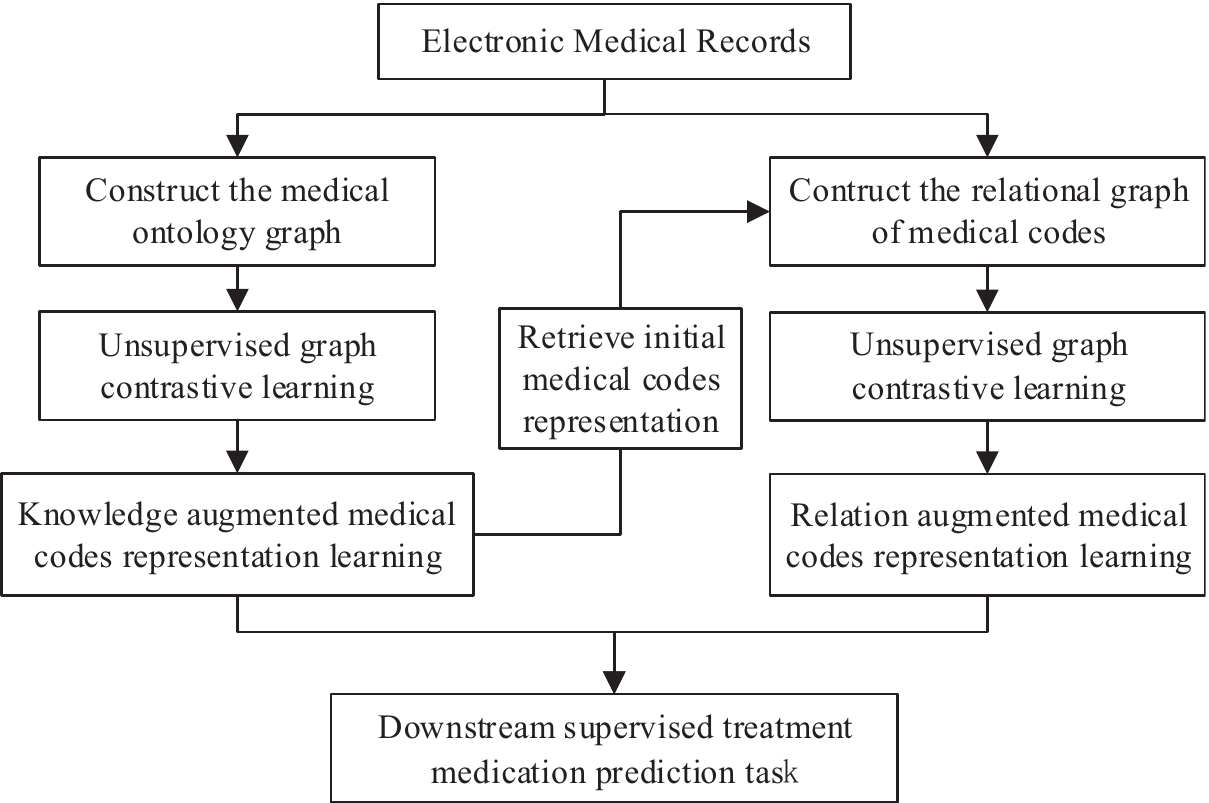}
    \caption{Overall framework of KnowAugNet.}
    \vspace{-0.2cm}
    \label{fig:Chap_BasicFramework}
\end{figure}

As shown in Figure \ref{fig:Chap_BasicFramework}, the proposed KnowAugNet consists of two critical parts: the unsupervised medical codes representation learning part and the downstream supervised sequential medication prediction part. The former can be further divided into two other stages.
In the first stage, the medical ontology graph is constructed based on the medical codes in historical EMRs and combined with the corresponding hierarchical medical ontology. Then the unsupervised graph contrastive learning method is incorporated to compute the knowledge augmented medical codes representation vectors (Figure \ref{fig:Chap_Model_HGMIEN}).
In the second stage, the medical prior relation graph containing various relations between medical codes is first constructed based on the visit-level diagnosis codes and medication codes from historical EMRs according to the process shown in Figure \ref{fig:Chap_RG}. Then, the corresponding medical codes representation vectors calculated by the first stage are retrieved as the initial embedding vectors of the nodes (medical code) of the medical prior relation graph, and another similar unsupervised contrastive learning but with different graph encoder is introduced to obtain the relation augmented medical codes representations (Figure \ref{fig:Chap_Model_RGMIEN}).
Finally, in the downstream supervised sequential medication prediction task, the input corresponding medical codes representations in every timestamp of the sequential learning network can be retrieved directly from the obtained knowledge augmented medical codes representation set and relation augmented medical codes representation set, and then the temporal dependencies or relations between medical codes can be captured through the sequence learning network (Figure \ref{fig:Chap_Model_PredictionCN}).

\subsubsection{Knowledge augmented medical codes representation learning}
\label{sec:localMED}
\paragraph{Medical ontology graph construction}

The hierarchical structures of the diagnosis classification system ICD-9 and medication classification system ATC imply the meaningful relations between medical codes, which have been constructed as the medical domain knowledge in previous studies \cite{Choi2017GRAMGA,shang2019pre}. Similarly, the leaf nodes of the tree structure based graph comprise the medical codes of history EMRs, and the non-leaf nodes mainly come from the medical codes classification system ICD-9 and ATC. To consider the valuable information inherent in the ontology graph, Shang et al. \cite{shang2019pre} utilize a pre-training method from the field of natural language processing, which combines the graph node representation learning process with the BIRT \cite{devlin-etal-2019-bert} based pre-training method. However, the dependency on self-supervised labels results in the poor transferability of the learned medical codes representations, and the calculation consumption of BIRT based pre-training method is relatively large.
Thus, to avoid the dependency on the labels of graph representation learning and make the learned medical codes representations directly applied in the downstream tasks such as medication prediction, we incorporate the unsupervised graph contrastive learning method based on DGI\cite{velickovic2018deep} to learn the medical codes representations by maximizing the mutual information of graph representation for the first time.

\paragraph{Ontology graph contrastive learning}
\begin{figure}
\centering
    \includegraphics[width=1\linewidth]{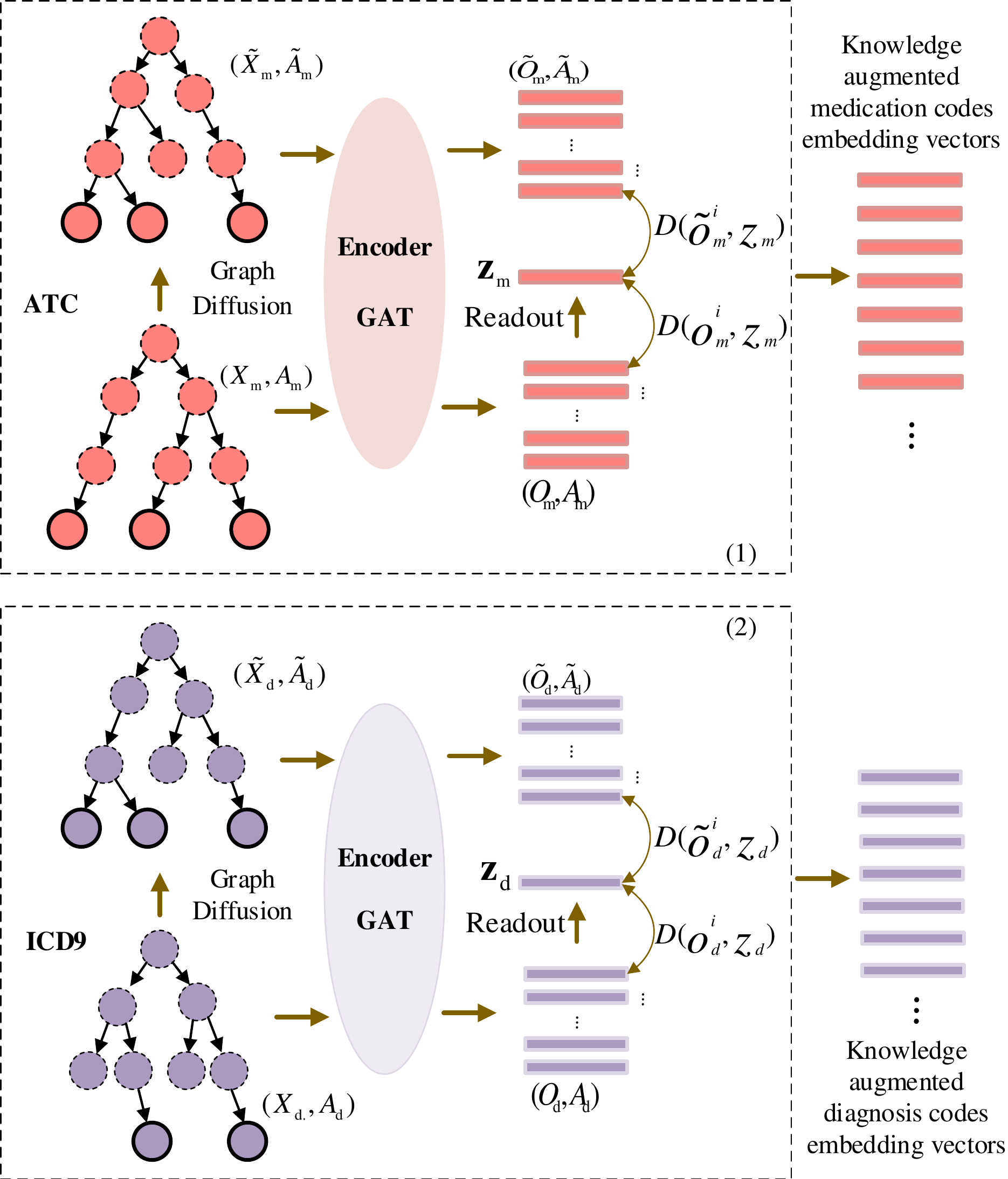}
    \caption{Contrastive learning on medical ontology graph.}
    \vspace{-0.2cm}
    \label{fig:Chap_Model_HGMIEN}
\end{figure}

As illustrated in Figure \ref{fig:Chap_Model_HGMIEN}, figure (1) and figure (2) respectively describe the introduced unsupervised contrastive learning framework for medical code representation learning of diagnosis code ontology graph and medication code ontology graph. Because of the similar critical hierarchical topology structure of the medical code ontology graphs (shown in Figure \ref{fig:Chap_HG}), in this section, we will just detail the graph contrastive learning process on diagnosis code ontology graph (Figure \ref{fig:Chap_Model_HGMIEN} (1)). Such a learning process consists of four critical substructures: graph data augmentation, GNN-based encoder, graph pooling layer, and the contrastive loss function.

Similar to previous knowledge-enhanced methods presented in G-BIRT \cite{devlin-etal-2019-bert}, the medical codes $c_i$ of ontology graph $\mathcal{G}_*$ can be randomly assigned an initialized embedding vector $\boldsymbol{v}_i$ which could be optimized and updated through a learned embedding matrix $\boldsymbol{W}_e\in\mathbb{R}^{\mathcal{C}_*\times{d}}$, where $d$ indicates the dimension of the medical code embedding vector.

\textbf{Graph data augmentation}. The constructed medical ontology graph $\mathcal{G}_* = \{\mathcal{V}_*,\mathcal{E}_*\}$ undergoes graph data augmentation to obtain the negative sample via corruption function such as the graph nodes permutation. In detail, the $\mathcal{C}$ is used to randomly perturb the nodes without changing the topology structure of the ontology graph; namely, the adjacency matrix $\mathcal{E}_*$, which aims to obtain a negative or pseudo ontology graph $\hat{\mathcal{G}}_*: (\hat{\mathcal{V}}_*,\hat{\mathcal{E}}_*)=\mathcal{C}(\mathcal{V}_*,\mathcal{E}_*)$.

\textbf{GNN-based encoder}. 
In fact, there should be no restrictions on the choice of graph encoders. Here, to fully utilize the inherent topological hierarchical structure of the medical ontology graph for further capturing the implicit relations between medical codes, we directly incorporate the graph attention network (GAT) \cite{Velickovic2018GraphAN} used in previous study G-BIRT \cite{devlin-etal-2019-bert} as the graph encoder $\mathcal{F}$ to obtain the medical codes embedding representations of medical ontology graphs.  

As illustrated in section \ref{PD}, the medical ontology graph $\mathcal{G}_*$ is a kind of hierarchical structural graph with a parent-child substructure. 
Based on this, the relation between medical codes can be captured from two different paths. That is, on the one hand, the medical codes corresponding to the parent nodes should infuse the information corresponding to the medical code of the child node; on the other hand, the embedding representation information corresponding to the medical code of the parent node also should be transmitted to the leaf node. For each non-leaf node in the medical domain knowledge graph $c^{'}\in\mathcal{C}^{'}$ and the leaf node $c_*\in\mathcal{C}_*$, the corresponding augmented embedding representations $\boldsymbol{v}_{c_*}\in \mathbb{R}^d$ and $\boldsymbol{h}_{o_*}\in \mathbb{R}^d$ can be computed as follows:
\begin{equation}
\centering
\label{Eq:Chapchild}
\begin{split}
\boldsymbol{v}_{c^{'}} &= f(c^{'},ch(c^{'}),\boldsymbol{W}_e),\\
\boldsymbol{o}_{c_*} &= f(c_*,pa(c_*),\boldsymbol{V}_e),
\end{split}
\end{equation}
where $f(\cdot,\cdot,\cdot)$ indicates the graph information aggregation function, $ch(c^{'})$ is a function to extract all direct child nodes of non-leaf medical code $c^{'}$, while $pa(c_*)$ represents a function which can extract all parent nodes of the leaf medical code $c_*$.
The above method realizes the two-way information transmission (from top to bottom and from bottom to top) through the hierarchical structure. In this way, the implicit relations between the medical codes can be fully captured; that is, the medical codes representations are augmented, which can further alleviate the sudden decline of prediction accuracy caused by the insufficient learning problems of the tail codes in electronic medical records.

Due to the graph information aggregation in above two-way information transmission process requires considering the relation difference between medical codes, in this paper, we incorporate the graph attention network (GAT) \cite{Velickovic2018GraphAN} as the aggregation function $f(\cdot,\cdot,\cdot)$, and it is also the shared graph encoder of the graph contrastive learning framework for calculating the graph nodes representations. Specifically, the representation vectors of each medical code $c_i$ is computed using the graph attention network aggregation function $f(\cdot,\cdot,\cdot) $ as follows:
\begin{equation}
\label{Eq:ChapGAT}
\begin{split}
f\left(c_{i}, p\left(c_{i}\right), \boldsymbol{H}_{e}\right)=\|_{k=1}^{K}
\sigma\left(\sum_{j \in \mathcal{N}_i} \alpha_{i, j}^{k} \boldsymbol{W}^{k} \boldsymbol{v}_{j}\right),
\end{split}
\end{equation}
where $\mathcal{N}_i = \{c_{i}\} \cup pa\left(c_{i}\right)$ denotes the first order neighborhood nodes set of medical code $c_i$ in the medical ontology graph, $\|$ is the concatenation operation of the embedding representations computed by multi-head attention, $K$ is the number of attention heads, $\sigma$ is the nonlinear activation function Sigmoid, $\boldsymbol{W}^{k}\in \mathbb{R}^{m \times d}$ is the transformation matrix to be learned, where $m=d/k$. While, $\alpha_{i, j}^{k}$ indicates the $k$-th standardized relevance score, which can be calculated as follows:
\begin{equation}
\label{Eq:ChapGATcof}
\begin{split}
\alpha_{i, j}^{k}=\frac{e^{\left(\operatorname{LeakyReLU}\left(\boldsymbol{a}^{\mathrm{T}}\left[\boldsymbol{W}^{k} \boldsymbol{v}_{i} \| \boldsymbol{W}^{k} \boldsymbol{v}_{j}\right]\right)\right)}}{\sum_{k \in \mathcal{N}_{i}} e^{\left(\operatorname{LeakyReLU}\left(\boldsymbol{a}^{\boldsymbol{\top}}\left[\boldsymbol{W}^{k} \boldsymbol{v}_{i} \| \boldsymbol{W}^{k} \boldsymbol{v}_{k}\right]\right)\right)}}
\end{split}
\end{equation}
where $\boldsymbol{a}\in \mathcal{R}^{2m}$ is the learned weight parameter, and $\operatorname{LeakyReLU}$ is the nonlinear activation function.

Therefore, we can utilize above graph encoder $\mathcal{F}$ to respectively compute the embedding representations of the medical codes from medical ontology graph $\mathcal{G}_*$ and pseudo medical ontology graph $\hat{\mathcal{G}}_*$ as follows:
\begin{equation}
\label{Eq:EmbOntoGraph}
\centering
\begin{split}
\boldsymbol{O}_{c_*} &= \mathcal{F}(\mathcal{V}_*,\mathcal{E}_*)=\{\boldsymbol{o}_{c_*^1},\boldsymbol{o}_{c_*^2},...,\boldsymbol{o}_{c_*^{|\mathcal{C}|}}\},\\
\hat{\boldsymbol{O}_{c_*}}&=\mathcal{F}(\hat{\mathcal{V}}_*,\hat{\mathcal{E}}_*)=\{\hat{\boldsymbol{o}}_{c_*^1},\hat{\boldsymbol{o}}_{c_*^2},...,\hat{\boldsymbol{o}}_{c_*^{|\mathcal{C}|}}\},
\end{split}
\end{equation}
where $\boldsymbol{O}_{c_*}$ and $\hat{\boldsymbol{O}_{c_*}}$ respectively denotes the embedding representations sets of medical codes from $\mathcal{G}_*$ and $\hat{\mathcal{G}}_*$.

\textbf{Graph pooling layer}. The graph pooling layer is mainly leveraged to compute the global feature vector $\boldsymbol{z}_*$ through $\operatorname{readout}$ function $\mathcal{R}$:
\begin{equation}
\label{Eq:ChapReadout1}
\centering
\begin{split}
\boldsymbol{z}_*&=\mathcal{R}(\boldsymbol{O}_{c_*})=\frac{1}{|\mathcal{C}_*|}\sum_{i=1}^{|\mathcal{C}_*|}\boldsymbol{o}_{c_*^i},
\end{split}
\end{equation}
where $\boldsymbol{O}$ indicates the unified symbol of embedding representation of medical codes of the medical ontology graph (diagnosis and medication).

\textbf{Contrastive loss function}. In order to train the graph encoder end-to-end and learn to obtain the informative medical codes embedding vectors and the medical ontology graph embedding representation, we still utilize the maximization of mutual information \cite{Hjelm2019LearningDR} between medical code embedding vector $\boldsymbol{o}_{c_*^i}$ and the medical ontology graph embedding representation $\boldsymbol{z}_*$ as the objective loss function.
First, the negative and positive sample pairs $(\boldsymbol{o}_{c_*^i},\boldsymbol{z}_*)$ and $(\hat{\boldsymbol{o}}_{c_*^i},\boldsymbol{z}_*)$ can be obtained through graph encoder and graph pooling layer. 
Then, the discriminator $\mathcal{D}$ is introduced to score the positive and negative sample pairs:
\begin{equation}
\label{Eq:ChapDiscriminate2}
\begin{split}
\mathcal{D}(\boldsymbol{o}_{c_*^i},\boldsymbol{z}_*)&=\sigma({\boldsymbol{z}_*}^T\boldsymbol{W}_{D}\boldsymbol{o}_{c_*^i} ),\\
\mathcal{D}(\hat{\boldsymbol{o}}_{c_*^i},\boldsymbol{z}_*)&=\sigma({\boldsymbol{z}_*}^T\boldsymbol{W}_{D}\hat{\boldsymbol{o}}_{c_*^i}).
\end{split}
\end{equation}
Finally, the overall objective function is to maximize the mutual information in the form of JS divergence as follows:
\begin{equation}
\label{Eq:ChapJSMIKG}
\begin{split}
\mathcal{L}=\frac{1}{|\mathcal{C}_*|}\sum_{i=1}^{|\mathcal{C}_*|} \mathbb{E}_{(\mathcal{V}, \mathcal{E})}\left[\log \mathcal{D}\left(\boldsymbol{o}_{c_*^i},\boldsymbol{z}_*\right)\right]+\\
    \frac{1}{|\mathcal{C}_*|}\sum_{j=1}^{|\mathcal{C}_*|} \mathbb{E}_{(\tilde{\mathcal{V}}, \tilde{\mathcal{E}})}\left[\log \left(1-\mathcal{D}\left(\hat{\boldsymbol{o}}_{c_*^i},\boldsymbol{z}_*\right)\right)\right]
\end{split}
\end{equation}

Through the above detailed graph contrastive learning on medical ontology graphs, we can obtain the knowledge augmented medical codes embedding vectors $\boldsymbol{O}_{c}$ including knowledge augmented medication codes embedding vectors $\boldsymbol{O}_{c_m}$ and knowledge augmented diagnosis codes embedding vectors $\boldsymbol{O}_{c_d}$ (as shown in Figure \ref{fig:Chap_Model_HGMIEN}).

\subsubsection{Relation augmented medical code representation learning}
\label{sec:GlobalRGMI}
In the actual clinical application scenario, patients will generally be diagnosed with a variety of diseases and given a variety of treatment medications during a single medical treatment, which also indirectly illustrates that there exist specific relations between diseases, medications, and between diseases and medications in the electronic medical records data.

Therefore, as shown in Figure \ref{fig:Chap_BasicFramework}, based on the history EMRs data generated by patients during each visit, and inspired by the dynamic weighted graph built on the history purchase records \cite{Yu2020PredictingTS}, we will construct the medical prior relation graph based on the co-occurrent medical codes including diagnosis codes and medication codes, and then embed the implicit topology relation information between medical codes into the embedding representation through incorporated unsupervised contrastive learning framework described in section \ref{sec:localMED}. Afterwards, we can obtain the relation augmented medical codes embedding vectors for the downstream predictive tasks such as medication prediction.

\paragraph{Medical prior relation graph construction}

\begin{figure*}
\centering
    \includegraphics[width=0.6\linewidth]{
    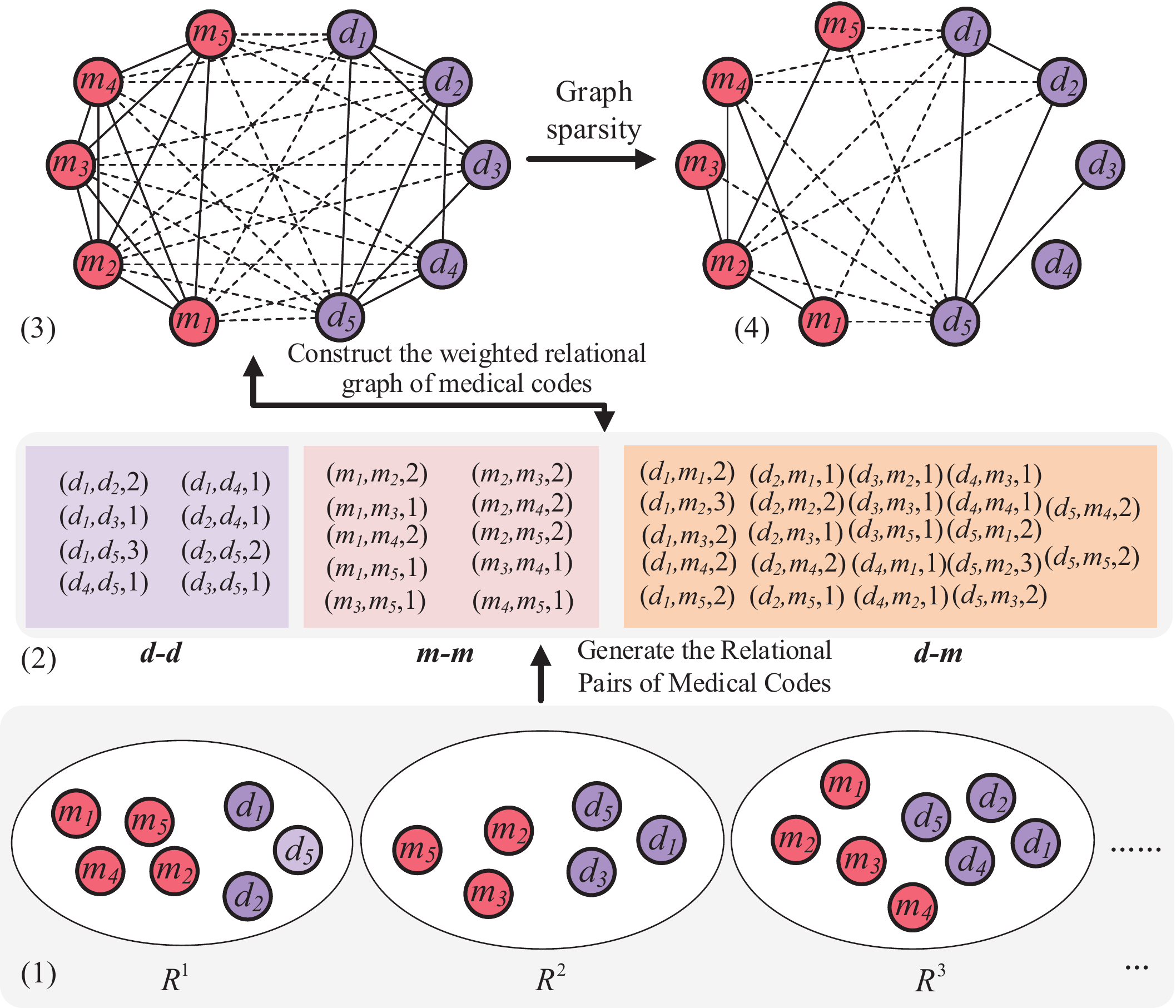}
    \caption{Construction process of prior medical knowledge graph.}
    \vspace{-0.2cm}
    \label{fig:Chap_RG}
\end{figure*}

Figure \ref{fig:Chap_RG} shows the detailed process of the medical prior relation graph construction based on the history EMRs of patients.
Since the patient with a single hospitalized visit has only one visit record, and the patient with multiple hospitalized visits has multiple visit records, in this paper, the construction of the medical prior knowledge graph does not consider the temporal sequence information but only considers the implicit relation between medical codes of history EMRs data.
Therefore, the visit-based medical record extracted from the patient's history EMRs is represented as $\boldsymbol{R}^{i}$, and every hospitalized visit could produce a diagnosis code and medication code.
As shown in Figure \ref{fig:Chap_RG}, there are three historical records: $\boldsymbol{R}^{1},\boldsymbol{R}^{2},\boldsymbol{R}^{3}$, and each record $\boldsymbol{R}^{i}$ includes diagnosis code $m_*$ and medication code $d_*$.
Based on the visit records shown in Figure \ref{fig:Chap_RG} (1), the corresponding medical code pairs in each visit record can be generated such as $(d_1,d_2)$, $(d_1,m_1)$, $(m_1,m_2)$, $\dots$. 
After mathematical statistics, the generated medical code pairs and the number of co-occurrence are shown in Figure \ref{fig:Chap_RG} (2). There are three implicit relations: the concurrent relation between diseases $d-d$, the synergistic relation between medications $m-m$, and the therapeutic relation between diseases and medication $d-m$.
Then, considering that the relations between medical codes are not only related to the number of co-occurrence but also related to the frequency of medical codes, the pointwise mutual information (PMI) \cite{Church1989WordAN} commonly used in natural language processing to measure the relevance of words is introduced to calculate the relation weights between medical codes:
\begin{equation}
\label{Eq:ChapPMI}
\begin{split}
PMI(c_i,c_j)&=\log\frac{P(c_i,c_j)}{P(c_i)P(c_j)}\\
            &=\log\frac{\frac{p(c_i,c_j)}{|\boldsymbol{R}|}}{\frac{p(c_i)}{|\boldsymbol{R}|}\frac{p(c_j)}{|\boldsymbol{R}|}}\\
            &=\log\frac{p(c_i,c_j)}{p(c_i)p(c_j)}|\boldsymbol{R}|,
\end{split}
\end{equation}
where $|\boldsymbol{R}|$ denotes the total number of visit records of all patients in the training dataset, $P(c_i,c_j)$ and $p(c_i,c_j)$ respectively denotes the probability and number of co-occurrence of medical codes $c_i$ and $c_j$ in single visit record, $P(c_i)$ and $p(c_i)$ indicates the probability and number of occurrence of medical code $c_i$ in visit records, $P(c_j)$ and $p(c_j)$ indicates the probability and number of occurrence of medical code $c_j$ in visit records.
The mutual information between the above medical code pairs is for all the code pairs, including $d-d$,$m-m$ and $d-m$ forms, without distinguishing whether they are homogeneous medical codes or heterogeneous medical codes. 
The main reason is that the relation degree between the medical code pairs is obtained by calculating statistical mutual information, which belongs to the scope of quantitative analysis without involving any medical background. Therefore, the qualitative relation is ignored in the quantitative calculation, even though the heterogeneity between medical codes does exist in reality.

Therefore, the medical prior relation graph $\mathcal{G}_r = \{\mathcal{V}_r,\mathcal{E}_r,\mathcal{W}_r\}$ as shown in Figure \ref{fig:Chap_RG} (3) can be built through the medical code pairs obtained in Figure \ref{fig:Chap_RG} (2) and the relation weights $PMI(c_i,c_j)$ computed in equation \ref{Eq:ChapPMI}.
The nodes in the medical prior relation graph are the set of medication codes and diagnosis codes, i.e. $\mathcal{V}_r=\mathcal{C}$. 
The medical code pairs obtained in Figure \ref{fig:Chap_RG} (2) determine the edges between nodes in the relation graph. Thus, the edge weights in adjacency matrix $\boldsymbol{A}$ of the medical prior relation graph $\mathcal{G}_r$ can be calculated as follows:
\begin{equation}
\label{Eq:ChapADJ}
\begin{split}
A(c_i,c_j)&=
\begin{cases}
   PMI(c_i,c_j) &\text{if } PMI(c_i,c_j)> \zeta, \\
   0 &\text{else}.
\end{cases}
\end{split}
\end{equation}
Different from the calculation method in \cite{su2020gate}, here, we incorporate a graph sparsity factor $0<\zeta<PMI_{max}$ in Equation (\ref{Eq:ChapADJ}), which aims to mitigate the effects of noise that might be introduced by relying solely on statistical quantitative computation method and ignoring medical expertise.
Moreover, the ralation graph belongs to a symmetric matrix, e.i. $A(c_j,c_i)=A(c_i,c_j)$.
In this way, the ultimately complete medical prior relation graph is obtained as shown in Figure \ref{fig:Chap_RG} (4)
In the end, another initial representation method of medical prior relation graph $\mathcal{G}_r$ can be obtained, e.i. $\mathcal{G}_r=(\mathcal{C},\boldsymbol{A})$.

\paragraph{Relation graph contrastive learning}
\begin{figure*}
\centering
    \includegraphics[width=0.8\linewidth]{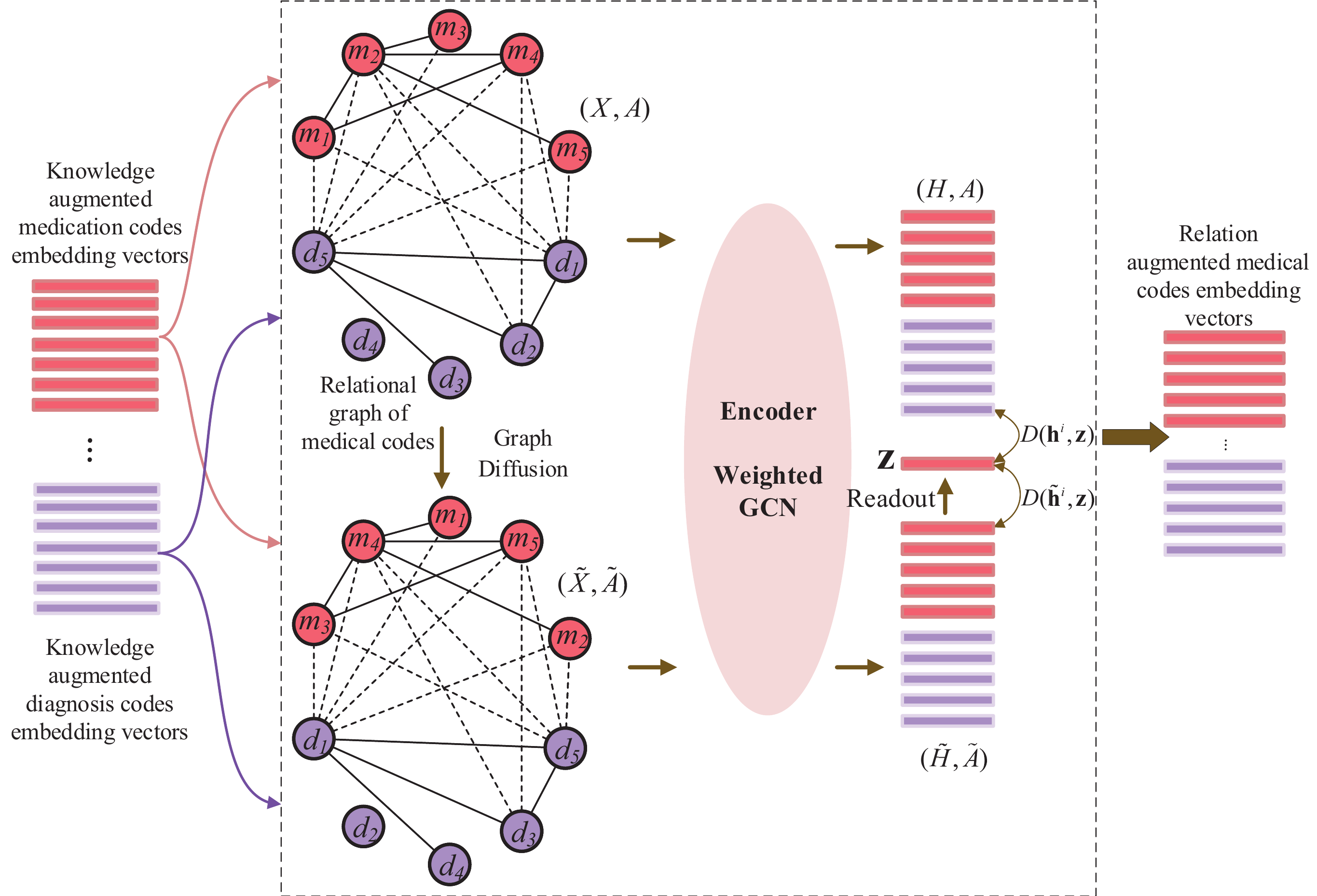}
    \caption{Contrastive learning on medical prior relation graph.}
    \label{fig:Chap_Model_RGMIEN}
\end{figure*}

In the previous section, based on the medical ontology graph, the knowledge augmented medical codes embedding vectors are obtained using the graph contrastive learning. Factually, the medical codes embedding vectors have infused the relational information from correlative medical codes, which can provide the initial embedding vectors for the nodes of the medical prior relation graph in this section.
However, the medical ontology graphs do not explicitly provide the practically meaningful relations between medical codes and the implicit relations between homogeneous medical codes in the hierarchical topology structure (belonging to a local relation).
The medical prior relation graph constructed based on history EMRs, in this section, directly models the explicit relations (belonging to a global relation) between homogeneous and heterogeneous medical codes through quantitative statistics and then uses the graph contrastive learning framework shown in Figure \ref{fig:Chap_Model_RGMIEN} to learn the relation augmented embedding vectors of the medical codes of the medical prior relation graph. 
The learning process will fully leverage the global relation between medical codes to model the interrelations of the medical codes embedding vectors.

As shown in Figure \ref{fig:Chap_Model_RGMIEN}, the unsupervised contrastive learning on medical prior relation graph also includes four critical steps.
First, the initialization vector of the corresponding node of medical codes from the medical prior relation graph can be retrieved directly from the knowledge augmented medical code embedding vectors set $\boldsymbol{O}_{c}$ obtained in section \ref{sec:localMED}, and denoted as $\boldsymbol{X}\subset\boldsymbol{O}_{c}$.
Then, the medical prior relation graph can be indicated as $\mathcal{G}_r=(\mathcal{C},\boldsymbol{A}):(\boldsymbol{X},\boldsymbol{A})$.
Subsequently, we can obtain the negative graph sample through graph nodes perturbation, namely pseudo medical prior relation graph $(\widetilde{\boldsymbol{X}},\widetilde{\boldsymbol{A}})\backsim(\boldsymbol{X},\boldsymbol{A})$.
After that, the incorporated shared graph encoder is incorporated to encode the above two relation graphs.
Different from the selected graph encoder in section \ref{sec:localMED}, the specially weighted graph convolutional network (GCN) \cite{Kipf2017SemiSupervisedCW} is incorporated as the graph encoder in the graph contrastive learning framework in this section to infuse the relation weights between medical codes. Taking the medical prior relation graph $(\boldsymbol{X},\boldsymbol{A})$ as an example, we calculate the corresponding embedding vectors as follows:
\begin{equation}
\label{Eq:ChapGCN}
\begin{split}
\boldsymbol{H}=\hat{\boldsymbol{D}}^{-1/2}\hat{\boldsymbol{A}}\hat{\boldsymbol{D}}^{-1/2}\boldsymbol{X}\Theta
\end{split}
\end{equation}
where $\hat{\boldsymbol{A}}=\boldsymbol{A}+\boldsymbol{I}$,$\boldsymbol{I}$ is the identity matrix which aims to avoid the information loss caused by the small number of neighbourhood nodes.
The diagonal matrix $\hat{\boldsymbol{D}}_{ii} = \sum_{j=0}A_{ij}$ is used to normalize the weights of the connected edges of each node from the relation graph according to the edge weights from the adjacency matrix $\hat{\boldsymbol{A}}$.
$\Theta\in\mathbb{R}^{d_x \times d_h}$ is the learned network parameter.
Considering the edge weights, the weighted graph encoder realizes the mutual infusion of correlative medical codes information in the medical codes embedding vectors learning process according to the relevance degree.
Similarly, with the help of above graph encoder weighted GCN, we can further encode the negative graph sample, e.i. pseudo medical prior relation graph $(\widetilde{\boldsymbol{X}},\widetilde{\boldsymbol{A}})$, and obtain the corresponding medical codes embedding matrix $(\widetilde{\boldsymbol{H}},\widetilde{\boldsymbol{A}})$ (shown in Figure \ref{fig:Chap_Model_RGMIEN}).
Afterwards, the $\operatorname{readout}$ function $\hat{\mathcal{R}}$ is used to compute the global embedding vector $\boldsymbol{z}$ of medical prior relation graph $\boldsymbol{H}={h_1,h_2,\dots,h_{|\mathcal{C}|}}$:
\begin{equation}
\label{Eq:ChapReadout2}
\begin{split}
\boldsymbol{z}&=\hat{\mathcal{R}}(\boldsymbol{O}_{c_*})=\frac{1}{|\mathcal{C}|}\sum_{i=1}^{|\mathcal{C}|}\boldsymbol{h}_i
\end{split}
\end{equation}

Finally, the constructed objective function based on the discriminant equation is introduced to optimize the graph contrastive learning process for the medical prior relation graph.
Similar to the contrastive learning framework for medical ontology graph, we still utilize the maximization of mutual information \cite{Hjelm2019LearningDR} between the embedding representation vector $\boldsymbol{h}_i$ of medical code on the medical prior relation graph and the global relation graph embedding vector $\boldsymbol{z}$.
Firstly, the negative and positive sample pairs $(\boldsymbol{h}_{i},\boldsymbol{z})$ and $(\hat{\boldsymbol{h}}_{i},\boldsymbol{z})$ can be obtained through graph encoder and graph pooling layer. 
Then, the discriminator $\mathcal{D}$ is introduced to score the positive and negative sample pairs:
\begin{equation}
\label{Eq:ChapDiscriminate1}
\begin{split}
\mathcal{D}(\boldsymbol{h}_{i},\boldsymbol{z})&=\sigma({\boldsymbol{z}}^T\boldsymbol{W}_{D}\boldsymbol{h}_{i} )\\
\mathcal{D}(\hat{\boldsymbol{h}}_{i},\boldsymbol{z})&=\sigma({\boldsymbol{z}}^T\boldsymbol{W}_{D}\hat{\boldsymbol{h}}_{i}).
\end{split}
\end{equation}
Then, the overall objective function is to maximize the mutual information in the form of JS divergence as follows:
\begin{equation}
\label{Eq:ChapJSMIRG}
\begin{split}
\mathcal{L}=\frac{1}{|\mathcal{C}|}\sum_{i=1}^{|\mathcal{C}|} \mathbb{E}_{(\boldsymbol{X}, \boldsymbol{A})}\left[\log \mathcal{D}\left(\boldsymbol{h}_{i},\boldsymbol{z}\right)\right]+\\
\frac{1}{|\mathcal{C}|}\sum_{j=1}^{|\mathcal{C}|} \mathbb{E}_{(\widetilde{\boldsymbol{X}}, \widetilde{\boldsymbol{A}})}\left[\log \left(1-\mathcal{D}\left(\hat{\boldsymbol{h}}_{i},\boldsymbol{z}\right)\right)\right].
\end{split}
\end{equation}
After continuous optimization and iterative calculation, the final relation augmented medical codes embedding representation vectors as shown in Figure \ref{fig:Chap_Model_RGMIEN} can be obtained, which, in the end, not only integrates the information from globally correlative heterogeneous and homogeneous medical codes embedding vectors in the medical prior relation graph but also infuses the information from locally correlative homogeneous medical codes embedding vectors implied in the medical domain knowledge graphs (the medical ontology graphs).

\subsubsection{Sequential learning network for medication prediction}
\label{sec:prediction}

\begin{figure*}
\centering
    \includegraphics[width=0.7\linewidth]{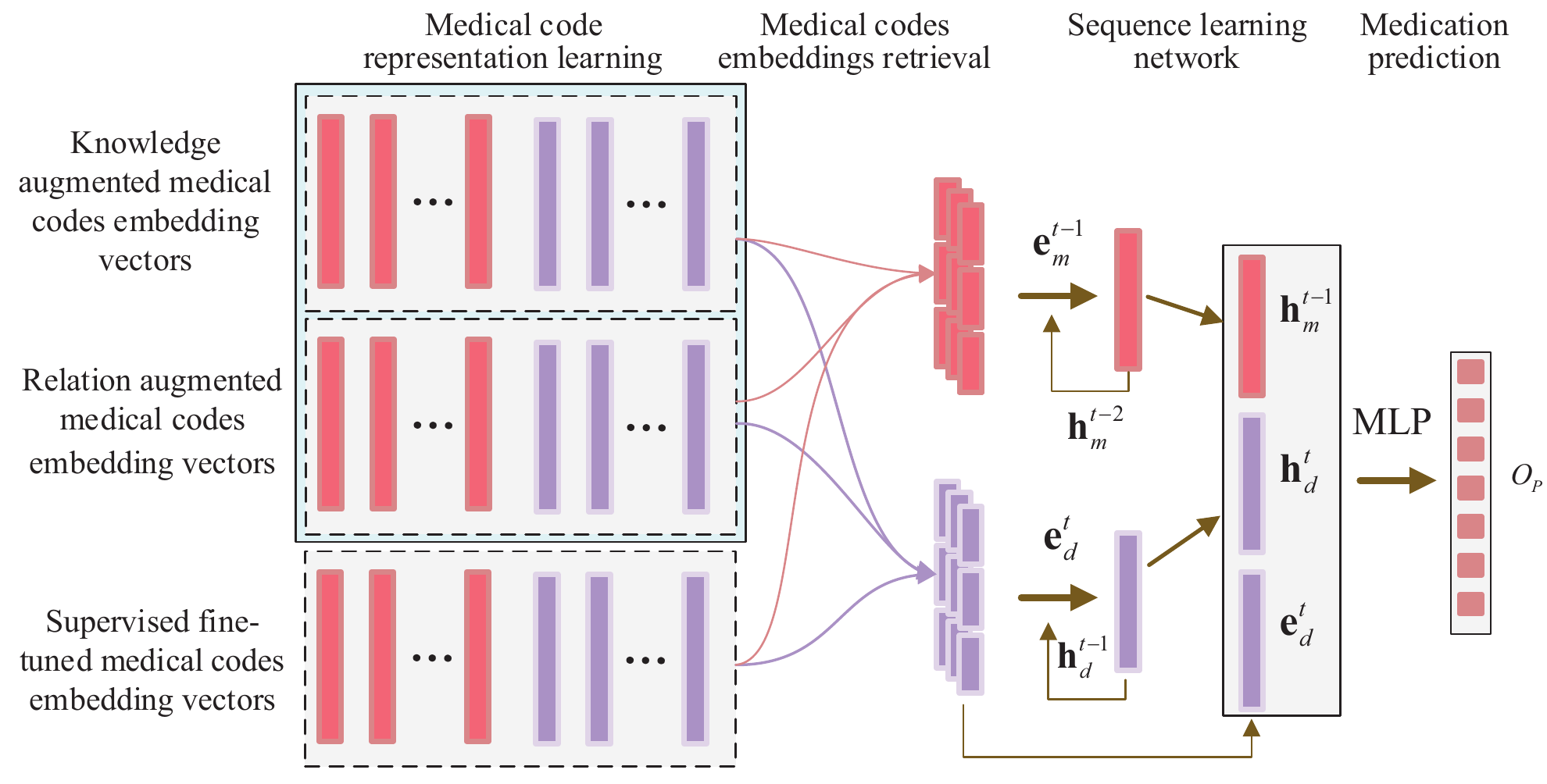}
    \caption{Medication prediction framework.}
    \vspace{-0.2cm}
    \label{fig:Chap_Model_PredictionCN}
\end{figure*}

As illustrated in Figure \ref{fig:Chap_Model_PredictionCN}, the sequential medication prediction framework comprise three critical substructures, e.i. multi-sourced medical codes are embedding vectors fusion, sequence learning network, and the comprehensive medication prediction.

\textbf{Multi-sourced medical codes embedding vectors fusion}. The knowledge augmented medical codes embedding vectors $\boldsymbol{O}_{c} = \{\boldsymbol{O}_{c_m},\boldsymbol{O}_{c_d}\}$ and the relation augmented medical codes embedding vectors $\boldsymbol{H}$ are respectively obtained through corresponding graph contrastive learning framework in section \ref{sec:localMED} and section \ref{sec:GlobalRGMI}.
While for each specific prediction task, it has different task-specific requirements for the medical codes embedding vectors. Therefore, Therefore, in the medication prediction task, the learned medical codes embedding representation vector $\boldsymbol{e}^i\in \mathbb{R}^d$ that can be supervised and fine-tuned in the specific prediction task is also incorporated:
\begin{equation}
\centering
  \label{Eq:ChapEmb}
  \boldsymbol{e}^i = \boldsymbol{W}^{e}\boldsymbol{c}^i,
\end{equation}
where $\boldsymbol{W}^{e} \in\mathbb{R}^{|\mathcal{C}|\times d}$ is the medical codes embedding matrix to be learned. In this way, we can get the dense embedding vector matrix of medical codes that can be learned, i.e. $\boldsymbol{E} = [\boldsymbol{e}^1,\dots,\boldsymbol{e}^{|\mathcal{C}|}] \in \mathbb{R}^{|\mathcal{C}|\times d}$.

Each hospitalized visit will generate different medical codes corresponding to different diagnoses and medications. The corresponding medical codes embedding vectors could be retrieved from the obtained medical codes embedding vectors sets, including knowledge augmented medical codes embedding vectors set $\boldsymbol{O}_{c}$, relation augmented medical codes embedding vectors set $\boldsymbol{H}$, and the supervised learning medical codes embedding vectors set $\boldsymbol{E}$.
For instance, we can first retrieve the corresponding embedding representations of medical codes produced in $i$-th visit record from the medical codes embedding vectors sets, i.e. the medical diagnosis codes embedding representations including $\boldsymbol{O}^i_{d}\subset\boldsymbol{O}_{c}$, $\boldsymbol{H}^i_{d}\subset\boldsymbol{H}$ and $\boldsymbol{E}^i_{d}\subset\boldsymbol{E}$, the medical medication codes embedding representations including $\boldsymbol{O}^i_{m}\subset\boldsymbol{O}_{c}$, $\boldsymbol{H}^i_{m}\subset\boldsymbol{H}$ and $\boldsymbol{E}^i_{m}\subset\boldsymbol{E}$. Then the corresponding mean values of medical codes representation vectors sets are calculated to obtain the visit-level diagnosis codes embedding vectors containing $\boldsymbol{o}^i_d$, $\boldsymbol{h}^i_d$ and $\boldsymbol{e}^i_d$, and medication codes embedding vectors containing $\boldsymbol{o}^i_m$, $\boldsymbol{h}^i_m$ and $\boldsymbol{e}^i_m$; Finally, the calculated embedding vectors are respectively concatenated together to obtain the input of diagnosis codes sequence learning network and medication sequence learning network, i.e. $\boldsymbol{x}^i_d = [\boldsymbol{o}^i_d,\boldsymbol{h}^i_d,\boldsymbol{e}^i_d]$ and $\boldsymbol{x}^i_m = [\boldsymbol{o}^i_m,\boldsymbol{h}^i_m,\boldsymbol{e}^i_m]$.

\textbf{Sequence learning network}. When patient possesses multiple hospitalized visit records and needs to predict the treatment medications at current timestamp $t$, firstly, we require integrating described multi-sourced medical codes embedding vectors together at history timestamps, including the history diagnosis codes embedding representation sequence $\{\boldsymbol{x}^1_d,\boldsymbol{x}^2_d,\dots,\boldsymbol{x}^t_d\}$ and the history medication codes embedding representation sequence $\{\boldsymbol{x}^1_m,\boldsymbol{x}^2_m,\dots,\boldsymbol{x}^(t-1)_m\}$; afterwards, the temporal sequential learning network such as the recurrent neural networks (RNNs) are respectively utilized to capture the temporal dependencies of sequential medical codes as follows:
\begin{equation}
 \label{Eq:ChapRNN}
 \begin{split}
 \boldsymbol{h}^t_m &= \boldsymbol{RNN}_{m}(\boldsymbol{x}^1_m,\boldsymbol{x}^2_m,\dots,\boldsymbol{x}^{t-1}_m),\\
\boldsymbol{h}^t_d &= \boldsymbol{RNN}_{d}(\boldsymbol{x}^1_d,\boldsymbol{x}^2_d,\dots,\boldsymbol{x}^t_d).
 \end{split}
\end{equation}

\textbf{Medication prediction}.
The hidden state vectors $\boldsymbol{h}^t_m$ and $\boldsymbol{h}^t_d$ that incorporates the history information are obtained through the sequence learning network (Eq. (\ref{Eq:ChapRNN})).
However, considering the importance of current diagnosis information for the medication prediction at the current timestamp, the current diagnosis code embedding vector $\boldsymbol{x}^t_d$ is also integrated into the patient representation. Therefore, the comprehensive patient representation vector $\boldsymbol{O}_P$ can be calculated as follows:
\begin{equation}
 \label{Eq:ChapOp}
 \begin{split}
 \boldsymbol{O}_P = \boldsymbol{W}_P\cdot[\boldsymbol{h}^t_m,\boldsymbol{h}^t_d,\boldsymbol{x}^t_d],
 \end{split}
\end{equation}
where $\boldsymbol{W}_P\in\mathbb{R}^{{4d_e+d_o+d_h}\times{d_e}}$ is the parameter to be learned. According to the comprehensive patient representation vector $\boldsymbol{O}_P$, the current treatment medication $\hat{\boldsymbol{y}}^m_{t}$ can be predicted as follows:
\begin{equation}
\label{Eq:Chap_Yt}
    \hat{\boldsymbol{y}}^m_{t} = \operatorname{softmax}(\boldsymbol{W}_O\cdot\boldsymbol{O}_P+\boldsymbol{b}_{o}),
\end{equation}
where $\hat{\boldsymbol{y}}^m_{t}$ denotes the predicted multi-label medications set, $\boldsymbol{W}_O\in\mathbb{R}^{{|\mathcal{C}_{m}|}\times{d_e}}$  and $\boldsymbol{b}_{o}\in\mathbb{R}^{|\mathcal{C}_{m}|}$ are the parameters to be learned.

Due to the medication prediction task belonging to the domain of sequential multilabel prediction, we utilize the binary cross-entropy loss $\mathcal{L}$ as the objective function.
According to the prediction result $\hat{\boldsymbol{y}}^m_{t}$ at each timestamp $t$ and the real label $\boldsymbol{y}^m_{t}$, the predictive function binary cross-entropy loss is formulated as follows:
\begin{equation}
\label{Eq:loss}
\mathcal{L}=-{\frac 1 {T-1}}\sum_{t=2}^{T} \boldsymbol{y}^m_{t} \log \sigma\left(\hat{\boldsymbol{y}}^m_{t}\right)+\left(1-\boldsymbol{y}^m_{t}\right)\log\left(1-\sigma\left(\hat{\boldsymbol{y}}^m_{t}\right)\right)
\end{equation}%

\section{Experiments and discussion}
\label{EX}
This section gives a quantitative evaluation of the performance of KnowAugNet.
First, the experiment details are introduced, including the selection and preprocessing of the experimental dataset, the evaluation metrics, the benchmark models and the experimental setting. Then, the experimental results of the overall performance are discussed, which contain the discussion of prediction results, the ablation study on model components, the influence of graph sparsity factor in medical prior relation graph, and the effects of graph encoders in graph contrastive learning frameworks. Finally, the limitations and future work are also discussed in the paper.

\subsection{Experimental details}
\label{sec:EXD}

\subsubsection{Datasets description}
\label{data}
\renewcommand\arraystretch{1.4}
\begin{table}
  \caption{Statistics of the MIMIC-III datasets}
  \vspace{0.2cm}
  \centering
  \begin{tabular}{l|c}
  \hline
  MIMIC III           & \multicolumn{1}{c}{Quantity} \\ \hline
  \# of patients (Single visit)                   & 30745 \\
  avg \# of diagnosis      & 39 \\
  avg \# of medication     & 52 \\
  \# of unique diagnosis      & 1997 \\
  \# of unique medication     & 323 \\\hline
  \# of patients (Multi visit)                   & 6350 \\
  avg \# of diagnosis      & 10.16 \\
  avg \# of medication     & 7.33 \\
  avg \# of visits         & 2.36 \\
  \# of unique diagnosis      & 1958 \\
  \# of unique medication     & 145 \\\hline
  \end{tabular}
  \label{tab:Chap_dataset}
\end{table}

As analyzed in Section \ref{Introduction}, KnowAugNet is proposed to fully mine the meaningful correlative relations between medical codes implicit in the multi-sourced medical knowledge for better the medications prediction performance. 
Hence, to validate the effectiveness of the proposed KnowAugNet, in this paper, we conduct the experiments on a publicly available dataset MIMIC-III \cite{MIMIC} \footnote{https://mimic.physionet.org}, in which patients stayed within the intensive care units (ICU) at the Beth Israel Deaconess Medical Center and had relatively complete multi-sourced EMRs. The patients in the experimental cohort have at least one complete visit record.
Furthermore, the patients with only one hospitalized visit record are mainly utilized to construct the medical knowledge graphs, including the ontology and relation graphs. At the same time, the patients with multiple hospitalized visit records are divided into train set and test set. The EMRs of patients in the train set is also used to assist in constructing the multi-sourced medical knowledge graph.
Besides, similar to \cite{Shang2019GAMENetGA}, the medications prescribed by doctors for each patient within the first 24 h are selected as the medication set since it belongs to a crucial period for each patient to get rapid and accurate treatment \cite{Fonarow2005Effect}. In addition, the medicine codes form NDC are transformed to ATC Level 3 for integration with MIMIC-III, and predicting category information not only guarantees the sufficient granularity of all the diagnoses but also improves the training speed, and predictive performance \cite{Ma2017DipoleDP,Choi2017GRAMGA}. 
Table \ref{tab:Chap_dataset} provides more information about the patient cohort from the dataset.

\subsubsection{Evaluation metrics}
\label{sec:metrics}
To evaluate the performance of the proposed KnowAugNet, the Jaccard similarity score (Jaccard), precision-recall AUC (PR-AUC), and average F1 (F1) are adopted as the evaluation metrics. Among them, Jaccard is defined as the size of the intersection divided by the size of the union of the predicted set $\hat{Y}_{t}^{m}$ and ground truth set $y_{t}^{m}$. 
Besides, for the medication prediction task, due to the number of positive and negative labels being imbalanced, the precision-recall curve utilized to calculate the PR-AUC has proved to be an appropriate evaluation metric.
While precision is generally adopted to measure predicted medications' correctness, the recall measures the predicted medications' completeness. Our proposed model, KnowAugNet cannot wholly replace doctors and only screen possible medications as much as possible to assist physicians in prescribing medications for patients. In addition, F1 is often introduced as a comprehensive evaluation metric for multilabel classification tasks.
\begin{equation}
\begin{split}
\text{Jaccard}=\frac{1}{\sum_{i}^{N} \sum_{t}^{T_{i}} 1} \sum_{i}^{N} \sum_{t}^{T_{i}} \frac{\left|Y_{t}^{i} \cap \hat{Y}_{t}^{i}\right|}{\left|Y_{t}^{i} \cup \hat{Y}_{t}^{i}\right|},
\end{split}
\end{equation}
where $T_{i}$ is the number of visits for the $i^{th}$ patient, and $N$ denotes the number of patients in the test set. Given
\begin{equation}
\begin{split}
\text{Recall}=\frac{1}{\sum_{i}^{N} \sum_{t}^{T_{i}} 1} \sum_{i}^{N} \sum_{t}^{T_{i}}
\frac{\left|Y_{t}^{i} \cap\hat{Y}_{t}^{i}\right|}
{\left|Y_{t}^{i}\right|}
\end{split},
\end{equation}
\begin{equation}
\begin{split}
\text{Precision}=\frac{1}{\sum_{i}^{N} \sum_{t}^{T_{i}} 1} \sum_{i}^{N} \sum_{t}^{T_{i}}
\frac{\left|Y_{t}^{i} \cap\hat{Y}_{t}^{i}\right|}
{\left|\hat{Y}_{t}^{i}\right|}
\end{split},
\end{equation}
the comprehensive evaluation metric F1 can be computed as follows:
\begin{equation}
\begin{split}
\text{Jaccard}=\frac{1}{\sum_{i}^{N} \sum_{t}^{T_{i}} 1} \sum_{i}^{N} \sum_{t}^{T_{i}} \frac{2\times \text{Precision}\times \text{Recall}}{\text{Precision} + \text{Recall}}.
\end{split}
\end{equation}

\subsubsection{Benchmark models}
To verify the superiority of the proposed model KnowAugNet on medication prediction tasks, we compare it with the following baseline methods used for medication prediction, which include one machine learning-based method and six deep learning-based algorithms:
\begin{itemize}[leftmargin=*]
    \item \textit{LR \cite{Luaces2012BinaryRE}}. It is a logistic regression with L1/L2 regularization. We sum the multi-hot vector of each visit together and apply the binary relevance technique \cite{Luaces2012BinaryRE} to handle the multilabel output.
    \item \textit{Retain \cite{Choi2016RETAINAI}}. RETAIN is an interpretable model with a two-level reverse time attention mechanism to predict diagnoses, which can detect significant past visits and associated clinical variables. It can be used for similar sequential prediction tasks, such as predicting treatment medicines.
    \item \textit{LEAP \cite{Zhang2017LEAPLT}}. Leap formulates the medicine prediction problem as a multi-instance multilabel learning problem, mainly using a recurrent neural network (RNN) to recommend medicines.
    \item \textit{GRAM\cite{Choi2017GRAMGA}}. It utilizes the diagnosis ontology graph to enhance the diagnosis code representation by infusing the information from relational medical codes in a supervised method.
    \item \textit{GAMENet \cite{Shang2019GAMENetGA}}. It employs a dynamic memory network to save encoded historical medication information, and further utilizes a query representation formed by encoding sequential diagnosis and procedure codes to retrieve medications from the memory bank.
    \item \textit{G-BIRT \cite{shang2019pre}}. G-BIRT combines the graph neural network and bidirectional encoder representation from transformers (BERT) to enhance medical code representations through a pre-training method.
    \item \textit{GATE \cite{su2020gate}}. GATE constructs the dynamic co-occurrence graph at each admission record for every patient. It then introduces a graph-attention augmented sequential network to model the inherent structural and temporal information for medication prediction. 
\end{itemize}

\subsubsection{Experimental settings}
\label{sec:settings}
To ensure the rationality of experimental verification, the pretreated clinical patients' EMRs data are still randomly divided into training, validation, and test sets with 2/3 : 1/6 : 1/6 ratios and the experimental results are the average values across five runs of random grouping and training. 
Moreover, the dimension of hidden layers and hyper-parameters in KnowAugNet is set as follows:
in the unsupervised graph contrastive learning framework for medical ontology graph, the dimension of hidden layers are set to 128, the node embedding size of graph encoder GAT is set to 32, the attention head is set to 4; in the unsupervised graph contrastive learning framework for medical prior relation graph, the dimension of hidden layers and the node embedding size of encoder GCN are all set to 64; the dimension of hidden layers in temporal sequential learning network is set to 256. 
In addition, the training is performed using Adam \cite{Kingma2014AdamAM} at a learning rate of 5e-4, and we report the model performance in the test set within 40 epochs.
All methods are trained on a Windows with 11GB memory and an Nvidia 2080Ti GPU using the deep learning computation platform Pytorch 1.6.

\subsection{Discussion}

\subsubsection{Discussion of prediction results}

\begin{table}
  \centering
  \caption{Performance comparison on medication prediction task}
  \begin{tabular}{l|ccc}
  \hline
  Methods                              & Jaccard     & F1   & PR-AUC \\ \hline
  $\operatorname{LR}$        &0.4075 &0.5658  &0.6716 \\
  $\operatorname{LEAP}$      &0.3921 &0.5508 &0.5855 \\
  $\operatorname{Retain}$    &0.4456 &0.6064 &0.6838 \\
  $\operatorname{GRAM}$      &0.4176 &0.5788 &0.6638 \\
  $\operatorname{GAMENet}^{-}$ &0.4401 &0.5996 &0.6672 \\
  $\operatorname{GAMENet}$  &0.4555 &0.6126 &0.6854 \\
  $\operatorname{G-BIRT}$    &0.4565 &0.6152 &0.6960 \\
  $\operatorname{GATE}$    &0.4742 &0.6315 &0.7087 \\\hline
  $\operatorname{KnowAugNet}$     & \textbf{0.4973} &\textbf{0.6454}  &\textbf{0.7195}   \\
  \hline
  \end{tabular}
  \label{tab:Chap_results}
\end{table}

As demonstrated in Table \ref{tab:Chap_results}, the treatment medication prediction results show that the performance of our proposed method KnowAugNet is better than the existing state-of-the-art predictive models in health informatics in most cases. 
In detail, compared with baseline model LEAP, KnowAugNet achieves about 10.52\%, 9.46\%, 13.4\% higher performance concerning Jaccard, F1, and PR-AUC, respectively.
We think the prominent reason might be that LEAP models the medication prediction problem as an instance-based medication prediction process which directly neglects the temporal dependency and does not consider the importance of implicit domain knowledge.
The medication prediction performance of Retain is relatively better than LEAP, which could be attributed to its two-level attention-based model, which can capture the temporal relations and the relation between input features and output labels. Such a two-level attention-based sequential prediction model makes Retain's performance even better than the GRAM model, which firstly introduces the hierarchical knowledge into the healthcare predictive model through the attention mechanism.

Secondly, the similarity between GAMENet and KnowAugNet is that the obtained diagnoses sequence and treatment medications sequence all utilize a sequence learning model to capture the temporal dependencies between medical codes. The difference is that GAMENet does not construct the relations between medical codes directly but constructs the medication-visit graph for capturing the indirect relations between medications for later retrieval using the attention mechanism. In other words, GAMENet considers modelling the co-occurrence relations between medications, while our KnowAugNet takes the multiple relations between heterogeneous or homogeneous medical codes into consideration.
Therefore, on the medication prediction task, our KnowAugNet outperforms GAMENet by 4.18\%, 3.28\%, 3.41\% on Jaccard, F1, PR-AUC, respectively.
Unlike GAMENet, G-BIRT does not directly construct the relation between medical codes. Instead, it enhances the relation between medical codes using a pre-training method on the medical ontology graphs, including diagnosis code ontology graph and medication code ontology graph, which can make full use of the EMRs data of single hospitalized patients. However, medical code embedding representation learning in G-BIRT mainly relies on the medication or diagnosis labels provided in history EMRs and ignores the inherent relations between medical codes implied in EMRs data.
Although GATE also considers the relations between medical codes by building the co-occurrence graphs for each patient from the global guidance relation graph, it still neglects the infusion of valuable information of correlative medical codes from the medical prior relation graph and relies on the supervised label of the medication prediction task. 
Therefore, KnowAugNet performs better than the latest knowledge-enhanced algorithm G-BIRT and GATE, and its Jaccard, F1 and PR-AUC improve at least by 2.31\% on Jaccard, by 1.39\% on F1 and by 1.08\% on PR-AUC, respectively.

The critical reason that our proposed KnowAugNet achieves the best performance compared with baseline models could be summarized as follows: (1) With the help of a multi-level unsupervised contrastive learning framework, it can capture the relations between medical codes and augments the medical codes representations based on the medical ontology graphs. (2) Then, the relations between medical codes implicit in the constructed medical prior relations graph are further captured to learn more informative medical codes embedding representation vectors, contributing to the downstream tasks such as the medication prediction. (3) The incorporated sequential learning network can further combine the supervised medical codes representations with the learned knowledge and relation augmented medical codes representations and then captures the temporal relations between medical codes for downstream medication prediction task.

\subsubsection{Ablation study on model components} 
\label{sec:ablation}
To verify the effectiveness of the critical components of KnowAugNet and analyze their influence on the performance of medication prediction tasks, we conduct an ablation study to explore further the necessity of the proposed model components in our multi-level graph contrastive learning framework on the multi-sourced medical knowledge for the medication prediction task.

\begin{table}
\centering
\caption{Performance Comparison of the Variants of KnowAugNet on MIMIC-III Dataset}
\begin{tabular}{l|ccc}
\hline
Model      & Jaccard   & F1  & PR-AUC\\\hline
  $\operatorname{KnowAugNet}_{{RG}^{-}}$ &0.4825	&0.6353	&0.7173\\
  $\operatorname{KnowAugNet}_{{HG}^{-}}$ &0.4792	&0.6313	&0.7144 \\
  $\operatorname{KnowAugNet}_{{HGRG}^{-}}$ &0.4809	&0.6325	&0.7138\\
  $\operatorname{KnowAugNet}_{R^{-}}$ &0.4866	&0.6357	&0.7147 \\
  $\operatorname{KnowAugNet}_{{RG}^{W-}}$ &0.4911	&0.6396	&0.7165 \\\hline
  $\operatorname{KnowAugNet}$ & \textbf{0.4973} &\textbf{0.6454}  &\textbf{0.7195}   \\
  \hline
\end{tabular}
\label{tab:Chap_Ablation}
\end{table}

As illustrated in Table \ref{tab:Chap_Ablation}, compared with KnowAugNet, the performance of five model variants decreased to varying degrees. We think the reason might be that the lack of model components leads to the failure of effective mining of the valuable relations between medical codes implicit in the multi-sourced medical knowledge. The details of the five variants are as follows:
\begin{itemize}[leftmargin=*]
    \item $\operatorname{KnowAugNet}_{{RG}^{-}}$.
    The variant $\operatorname{KnowAugNet}_{{RG}^{-}}$ does not consider the medical prior relation graph constructed based on the empirical knowledge from the history EMRs data. That is, the relation augmented medical codes embedding representation vectors $\boldsymbol{h}^i_d$ and $\boldsymbol{h}^i_m$ are respectively excluded from the input of the corresponding sequence learning network. the performance of variant $\operatorname{KnowAugNet}_{{RG}^{-}}$ decreases by 1.48\% on Jaccard, 1.01\% on F1, and 0.22\% on PR-AUC. The main reason is that the variant considers the locally inherent relations between medical codes in the medical ontology graph and does not consider capturing the valuable global relations between homogeneous or heterogeneous medical codes in the medical prior relation graph.
    
    \item $\operatorname{KnowAugNet}_{{HG}^{-}}$.
    The variant $\operatorname{KnowAugNet}_{{HG}^{-}}$ does not consider the medical domain knowledge graph, including the diagnosis code ontology graph and medication code ontology graph. That is, the knowledge augmented medical codes embedding representation vectors $\boldsymbol{o}^i_d$ and $\boldsymbol{o}^i_m$ are respectively excluded from the input of the corresponding sequence learning network.
    The performance of the variant declines by 1.81\% on Jaccard, 1.41\% on F1, 0.51\% on PR-AUC, respectively, which is mainly because the medical prior relation graph can not acquire the meaningful initialization vectors for its nodes from the medical ontology graph. Moreover, the valuable relations between medical codes embodied in the inherent hierarchical structures in medical ontology graphs would not be captured for augmenting the medical codes embedding vectors.
    
    \item $\operatorname{KnowAugNet}_{{HGRG}^{-}}$.
    The variant $\operatorname{KnowAugNet}_{{HGRG}^{-}}$ concurrently neglects the multi-sourced knowledge augmented medical codes embedding representation vectors and consider the supervised learnable medical codes embedding representations $\boldsymbol{e}^i_d$ and $\boldsymbol{e}^i_m$ in a sequence learning network. It achieves lower by 1.64\%, 1.29\%, 0.57\% than $\operatorname{KnowAugNet}$ respectively on Jaccard, F1, PR-AUC, but has relatively better performance than $\operatorname{KnowAugNet}_{{HG}^{-}}$, which demonstrates that there exists specific noise in the medical ontology graphs. In the future, we will explore how to effectively reduce the adverse effect of noise in multi-sourced medical knowledge.
   
    \item $\operatorname{KnowAugNet}_{R^{-}}$.
    The variant $\operatorname{KnowAugNet}_{R^{-}}$ indicates that the medical prior knowledge graph would not utilize the learned knowledge augmented medical codes embedding representation vectors in section \ref{sec:localMED} to initialize the graph nodes (medical codes) in section \ref{sec:GlobalRGMI}.
    It further validates the importance of such implicit information in the multi-level graph contrastive learning framework, which is proposed based on the relations between the medical ontology graph and the medical prior relation graph. It also indirectly explains why the performance of KnowAugNet is relatively optimal when using the knowledge augmented medical codes embedding representation vectors as the initialization vector of the nodes of the medical prior relation graph.
    
    \item $\operatorname{KnowAugNet}_{{RG}^{W-}}$.
    The variant $\operatorname{KnowAugNet}_{{RG}^{W-}}$ considers whether there are relations between medical codes and does not consider the relation weights reflected in the medical prior relation graph. In this way, the performance of variant $\operatorname{KnowAugNet}_{{RG}^{W-}}$ decreases compared with presented $\operatorname{KnowAugNet}$, which further testifies that the relation weights based on co-occurrence probability obtained via the statistical method in the medical prior relation graph have a positive promoting effect on the performance of medication prediction.
\end{itemize}

Therefore, the proposed KnowAugNet in this paper achieves the best performance only when the model's components complement each other. 

\subsubsection{Analysis on the graph sparsity factor $\zeta$}
\label{sec:factor}
In the construction process of the medical prior relation graph,  considering the adverse effects of the noise introduced by relying solely on the statistical quantitative computation method and ignoring medical expertise, the graph sparsity factor $0<\zeta<PMI_{max}$(Equation \ref{Eq:ChapADJ}) is incorporated in section \ref{sec:GlobalRGMI}. 
This section will explore the effect of graph sparsity factor $0<\zeta<PMI_{max}$ on medication prediction performance.

\begin{table}
  \centering
  \caption{The effect of graph sparsity factor $\zeta$ on model performance}
  \begin{tabular}{l|ccc}
  \hline
   $\zeta$      & Jaccard   & F1  & PR-AUC\\\hline
	0.01 &0.4917 &0.6398 &0.7176 \\
	0.02 &0.4908 &0.6392 &0.7163 \\
	0.03 &0.4895 &0.6384 &0.7187 \\
	0.04 &0.4948 &0.6431 &0.7199 \\
	0.05 &0.4916 &0.6407 &0.7176 \\
	0.06 &0.4924 &0.6407 &0.7196 \\
	0.07 &\textbf{0.4973} &\textbf{0.6454} &0.7195 \\
	0.08 &0.4922 &0.6404 &\textbf{0.7201} \\
	0.09 &0.4933 &0.6413 &0.7194 \\
	0.10 &0.4903 &0.6382 &0.7175 \\
  \hline  \end{tabular}
  \label{tab:Chap_Zeta}
\end{table}
Table \ref{tab:Chap_Zeta} shows the prediction results of treatment medications when the medical prior relation graph is constructed based on different graph sparsity factors.
It can be seen from the table that when $\zeta=0.07$, the prediction performance of model KnowAugNet is the best, and the performance decreases in varying degrees with the increase or decrease of $\zeta$ value.
When $\zeta<0.07$, the redundancy of relations between medical codes might result in relatively more noise, which leads to a decline in prediction performance; 
when $\zeta>0.07$, with the increasing the value of the graph sparsity factor, some meaningful edges with beneficial relation might be neglected due to the sparsity, which would bring about the incomplete captures of the relations between medical codes, and would further cause the decline of model performance.
In general, the value of the sparsity factor has a relatively small impact on the prediction results compared with the critical components studied in section \ref{sec:ablation}.
The prominent reason may be that the noise influence in the medical prior relation graph is relatively small, and the valuable relations between medical codes and the augmentation of the embedding representation vectors of medical codes are still dominant.
In addition, the final representation of medical code is concatenated by multi-source medical codes embedding representations. The incomplete learning of one source medical code embedding representation can not have a noticeable adverse impact on the prediction performance of the model KnowAugNet.
Factually, it further shows that the proposed KnowAugNet is more robust in predicting treatment medications.

\subsubsection{Effects of graph encoders in contrastive learning networks}
\label{sec:Encoder}

\begin{table*}
  \centering
  \caption{The effect on prediction performance of different graph encoders}
\begin{tabular}{c|cc|cc|ccc}
\hline
\multirow{2}{*}{Model} & \multicolumn{2}{c|}{Encoder in HG} & \multicolumn{2}{c|}{Encoder in RG} & \multicolumn{3}{c}{Prediction Performance} \\ \cline{2-8}
 & \multicolumn{1}{c}{GAT} & \multicolumn{1}{c|}{GCN} & \multicolumn{1}{c}{GAT} & GCN & \multicolumn{1}{c}{Jaccard} & \multicolumn{1}{c}{F1} & \multicolumn{1}{c}{PR-AUC} \\ \hline
$\operatorname{KnowAugNet}_{AC}$ & \ding{51} & \ding{55} & \ding{55} & \ding{51}&0.4973	&0.6454	&0.7195 \\
$\operatorname{KnowAugNet}_{AA}$ & \ding{51} & \ding{55} & \ding{51} & \ding{55}&0.4878 &0.6381 &0.7160  \\
$\operatorname{KnowAugNet}_{CC}$ & \ding{55} & \ding{51} & \ding{55} & \ding{51}&0.4869 &0.6373 &0.7176   \\
$\operatorname{KnowAugNet}_{CA}$ & \ding{55} & \ding{51} & \ding{51}& \ding{55} &0.4853 &0.6358 &0.7122  \\
\hline
\end{tabular}
  \label{tab:Chap_Encoder}
\end{table*}

The multi-level unsupervised contrastive learning framework described in section \ref{sec:Chap_Methods}, including the medical domain knowledge graph-based contrastive learning framework and the medical prior relation graph-based unsupervised contrastive learning framework, possesses respectively different graph encoders because of distinct reasons.
In this section, we will analyze the selection of different graph encoders in the graph contrastive learning framework and explore their effects on downstream tasks such as medication prediction results.
In table \ref{tab:Chap_Encoder}, “Encoder in HG” indicates the applied graph encoder in the contrastive learning framework based on the medical domain knowledge graph (HG), while “Encoder in RG” represents the applied graph encoder in the contrastive learning on the medical prior relation graph (RG).
$\surd$ denotes using the corresponding graph encoder, while $\times$ denotes not using the corresponding graph encoder.

$\operatorname{KnowAugNet}_{AC}$ indicates that the graph encoders in HG and RG based contrastive learning frameworks are graph attention network (GAT) and weighted graph convolutional network (GCN), respectively, which achieves the relatively optimal prediction result. 
While in $\operatorname{KnowAugNet}_{CC}$, the graph encoder in HG based contrastive learning framework is replaced with general GCN, which results in a decrease in the medication prediction performance. We think the main reason is that GAT can aggregate the information of correlative neighbourhood codes to the leaf codes according to the learned relevance scores between connected medical codes, while general GCN does not consider the relation weights between medical codes and aggregate the information of connected (or correlative) medical codes equally.
Compared with variant $\operatorname{KnowAugNet}_{AC}$, variant $\operatorname{KnowAugNet}_{AA}$ directly uses GAT as the graph encoder, which could relearn the relevance score and does not consider the relation weights representing the empirical knowledge from medical prior relation graph. The result is the decrease of prediction performance.
Using GAT as the graph encoder in the medical prior relation graph-based contrastive learning framework is not appropriate. We think the main reason is that the learned normalized relation score between medical codes in GAT belongs to an uncertain relevance score. In contrast, the empirical relation weight between medical codes computed based on the co-occurrence probability in the medical prior relation graph is a relatively specific relevance score.
Therefore, compared to the above other variants, $\operatorname{KnowAugNet}_{CA}$ performs worst on the medication prediction task. 


\section{Conclusion and future work}
In this study, we propose a multi-sourced medical knowledge augmented medication prediction network. 
Expressly, we incorporate a novel multi-level graph contrastive learning framework for fully capturing the valuable relations between medical codes implicit in the multi-sourced medical knowledge.
The framework firstly leverages two local graph contrastive learning on the medical ontology graphs to learn the knowledge augmented embedding vectors of diagnosis codes and medication codes, which factually have infused the information of correlative medical codes into each other in the learning process.
Then, the medical prior relation graph is constructed and utilized to learn the relation augmented medical codes embedding vectors using the graph contrastive learning framework, which aims to capture the global relations between homogeneous and heterogeneous medical codes.
Finally, the multi-channel sequence learning network is presented to capture the temporal relations between medical codes, by which we can get a comprehensive patient representation for downstream tasks such as medication prediction.
We evaluate the performance of the proposed KnowAugNet on a public real-world clinical dataset, and the experimental results show that our model achieves the best medication prediction performance against baseline models in terms of Jaccard, F1 score, and PR-AUC.

With the help of the presented framework, in the future, we can introduce more related medical domain knowledge, such as the medication-related molecular graph and the bipartite graph representing the adverse medication reactions. In addition, we would cultivate more advanced algorithms to better mine the insightful information implicit in the multi-sourced medical knowledge and explore how to determine the importance of multi-sourced medical knowledge automatically.

\section*{Acknowledgements}
Funding: This research was partially supported by the National Key R\&D Program of China (2018YFC0116800), National Natural Science Foundation of China (No. 61772110, 6217072188 and 71901011), CERNET Innovation Project (NGII20170711).

\section*{Declaration of Competing Interest} 
Authors declare that there is no conflict of interest.
\appendix

\printcredits

\bibliographystyle{cas-model2-names}

\bibliography{cas-refs}


\end{document}